\documentclass{article} 
\usepackage[final]{colm2026_conference}
\usepackage{microtype}
\usepackage{xcolor}
\usepackage{graphicx}
\usepackage{hyperref}
\usepackage{url}
\usepackage{booktabs}

\usepackage{lineno}

\definecolor{darkblue}{rgb}{0, 0, 0.5}
\hypersetup{colorlinks=true, citecolor=darkblue, linkcolor=darkblue, urlcolor=darkblue}

\usepackage{amsmath,amsfonts,amssymb}
\usepackage{amsthm}
\usepackage{array}
\usepackage{multirow}
\usepackage{algorithm}
\usepackage{algorithmic}
\usepackage{colortbl}
\usepackage{tikz}
\usepackage{pgfplots}
\pgfplotsset{compat=1.17}
\usetikzlibrary{shapes,arrows,positioning,fit,calc,backgrounds,shadows,decorations.pathreplacing,matrix,patterns}

\newcommand{\rev}[1]{\textcolor{blue}{#1}}
\definecolor{fixred}{rgb}{0.65,0,0.1}
\newcommand{\fixnew}[1]{\textcolor{fixred}{#1}}
\definecolor{fixgreen}{rgb}{0,0.42,0.24}
\newcommand{\fixv}[1]{\textcolor{fixgreen}{#1}}
\definecolor{fixpurple}{rgb}{0.55,0,0.55}
\newcommand{\fixw}[1]{\textcolor{fixpurple}{#1}}
\renewcommand{\rev}[1]{#1}      
\renewcommand{\fixnew}[1]{#1}   
\renewcommand{\fixv}[1]{#1}     
\renewcommand{\fixw}[1]{#1}     

\newtheorem{theorem}{Theorem}
\newtheorem{proposition}{Proposition}

\theoremstyle{definition}
\newtheorem{definition}{Definition}
\newtheorem{remark}{Remark}
\theoremstyle{plain}

\definecolor{craftblue}{RGB}{66,133,244}
\definecolor{craftgreen}{RGB}{52,168,83}
\definecolor{craftorange}{RGB}{251,188,5}
\definecolor{craftred}{RGB}{234,67,53}
\definecolor{craftpurple}{RGB}{142,68,173}
\definecolor{lightblue}{RGB}{230,240,255}
\definecolor{lightgreen}{RGB}{230,255,230}
\definecolor{lightorange}{RGB}{255,245,230}
\definecolor{brown}{RGB}{139,69,19}

\newcommand{\colred}[1]{\textcolor{craftred}{#1}}

\newcommand{\R}{\mathbb{R}}
\newcommand{\PP}{\mathbb{P}}
\newcommand{\norm}[1]{\left\|#1\right\|}
\newcommand{\fnorm}[1]{\left\|#1\right\|_F}

\title{LORA-CRAFT: Cross-layer Rank Adaptation via Frozen Tucker Decomposition of Pre-trained Attention Weights%
\thanks{\normalfont Code and logs  will be available: \url{https://github.com/Kasun-Dewage/Lora_Craft_2026}}}

\author{
Kasun Dewage, \; Marianna Pensky, \; Shankhadeep Mondal \; \& \; Suranadi De Silva \\
University of Central Florida \\
Orlando, FL 32816, USA \\
\texttt{\{KasunTharuka.Dewage, Marianna.Pensky\}@ucf.edu} \\
\texttt{\{shankhadeep.mondal, su966204\}@ucf.edu}
}

\begin{document}

\lhead{Published as a conference paper at COLM 2026}

\ifcolmsubmission
\linenumbers
\fi

\maketitle

\begin{abstract}
We introduce LoRA-CRAFT (\textbf{C}ross-layer \textbf{R}ank \textbf{A}daptation via \textbf{F}rozen \textbf{T}ucker), abbreviated CRAFT throughout, an extremely parameter-efficient fine-tuning (PEFT) method that applies Tucker tensor decomposition to pre-trained attention weight matrices stacked across transformer layers and trains only small square adaptation matrices on the resulting frozen Tucker factors. Existing tensor-based PEFT methods decompose \textit{gradient updates}: LoTR applies Tucker decomposition with shared factor matrices, while SuperLoRA groups and reshapes $\Delta W$ across layers before applying Tucker decomposition. Separately, methods such as PiSSA apply SVD to \textit{pre-trained weights} but operate independently per layer. CRAFT bridges these two lines of work: it performs full Tucker decomposition via Higher-Order SVD (HOSVD) directly on \textit{pre-trained weights} organized as cross-layer 3D tensors, freezes all resulting factors, and adapts the model through lightweight trainable transformations applied to each factor matrix. Experiments on the GLUE benchmark using RoBERTa-base and RoBERTa-large, as well as commonsense reasoning benchmarks using LLaMA2-7B and LLaMA3-8B, demonstrate that CRAFT achieves competitive performance with existing methods while requiring only \rev{\textbf{extremely low Tucker adaptation parameters}}. \fixw{On LLaMA3-8B, CRAFT} \rev{exceeds the average accuracy of LoRA} \textbf{using hundreds of times fewer parameters}\fixw{; on LLaMA2-7B the same holds at a $0.252$M budget}. Our results suggest that CRAFT's efficiency advantage grows with model scale, as the frozen Tucker factors better capture the richer cross-layer structure of larger pre-trained models.
\end{abstract}

\section{Introduction}

The transformer architecture~\citep{vaswani2017attention} has become the dominant paradigm in natural language processing, but adapting large pre-trained models to downstream tasks remains computationally prohibitive. Parameter-efficient fine-tuning (PEFT) methods address this challenge by training only a small subset of parameters while keeping the bulk of the model frozen.
LoRA~\citep{hu2022lora} pioneered the approach of decomposing weight updates into low-rank matrices, reducing trainable parameters from millions to thousands. However, LoRA and its variants treat each weight matrix independently, missing a crucial insight: attention mechanisms in transformers exhibit strong \textit{multi-way correlations} across layers. Recently, two complementary research directions have begun to exploit this observation.

The first line of work analyzes \textit{gradient updates} using tensor decomposition techniques.
LoTR~\citep{bershatsky2024lotr} represents gradient updates as a tensor and applies Tucker decomposition with shared factor matrices across layers. SuperLoRA~\citep{chen2024superlora} extends this by grouping and reshaping $\Delta W$ across multiple layers before applying Tucker decomposition. VeRA~\citep{kopiczko2024vera} shares random frozen matrices across layers, and LoRETTA~\citep{yang2024loretta} employs tensor-train decomposition for ultra-low-parameter fine-tuning.
The second approach decomposes the \textit{pre-trained weights themselves}. PiSSA~\citep{meng2024pissa} applies SVD to pre-trained weight matrices, initializing LoRA adapters with the principal singular values and vectors while freezing the residual. However, PiSSA operates independently on each layer's weight matrix, missing cross-layer correlations.

Before proceeding, we briefly clarify Tucker decomposition terminology. The \textit{Tucker decomposition} of an $N$th-order tensor expresses it as a core tensor multiplied by a factor matrix along each mode. A \textit{Tucker-2} decomposition retains factor matrices for only two of three modes; for instance, LoTR~\citep{bershatsky2024lotr} employs Tucker-2. CRAFT employs Tucker-3 (full Tucker decomposition for third-order tensors), compressing all three modes simultaneously.

We propose CRAFT, which unifies these two directions by performing tensor decomposition on pre-trained weights organized as cross-layer tensors. CRAFT differs from all prior approaches in two key ways:

\textbf{1.\ Cross-layer decomposition of pre-trained weights.} Like PiSSA, CRAFT decomposes pre-trained weights rather than gradient updates. But unlike PiSSA's per-layer matrix SVD, CRAFT stacks weights across layers into 3D tensors and applies full Tucker-3 decomposition,
simultaneously capturing cross-layer, output-dimension, and input-dimension patterns.

\textbf{2.\ Frozen factors with trainable adaptation matrices.} While SuperLoRA and LoTR train the Tucker factors or core tensors themselves, CRAFT freezes \textit{all} Tucker factors (including the core tensor) obtained from HOSVD and introduces small square adaptation matrices $J^{(1)}, J^{(2)}, J^{(3)}$ as the \textit{only} trainable parameters, initialized near identity. This yields a trainable parameter count independent of both model dimension $d$ and number of layers $N_L$, for fixed Tucker ranks.

Following the standard LoRA convention~\citep{hu2022lora}, CRAFT adapts Q and V projections. The choice of Q over K is motivated by gauge invariance of the bilinear form $QK^\top$ and empirical PCA analysis; details are provided in Appendix~\ref{sec:empirical}. \rev{This is our main configuration; we additionally report a variant, CRAFT$^\star$, in which all four projections $\{Q, K, V, O\}$ are adapted.}

{\bf Advantages of CRAFT.\ }
The CRAFT framework offers several distinct advantages: 
\textbf{(a)} With a {\bf small number of Tucker adaptation parameters}, which at fixed ranks is \textbf{independent} of both the model dimension $d$ and the number of layers $N_L$, it achieves accuracy comparable to methods with significantly larger parameter budgets.
\textbf{(b)} Because only $n_p(r_1^2+r_2^2+r_3^2)$ parameters carry gradients, the \textbf{gradient and optimizer-state memory} is negligible relative to LoRA or full fine-tuning. We do not claim a wall-clock speed-up. Under identical
LLaMA3-8B profiling conditions, CRAFT's smallest configuration
($0.018$M trainable parameters) was approximately $0.5\%$ faster than
LoRA ($27.3$M), the main $(32,128,128)$ configuration ($0.067$M) was
$1.4\%$ slower within the observed $3$--$4\%$ run-to-run
variation and the largest configuration ($0.252$M) was $6.8\%$
slower, whereas DoRA ($27.3$M) required approximately $71\%$ more time
per optimizer step than LoRA.

\textbf{(c)} The method is {\bf theoretically grounded} with approximation guarantees inherited from HOSVD. 
Specifically, while in the case of additive LoRA adaptations, choosing the ranks of adaptations is a 
separate and rather cumbersome task, CRAFT preserves the effective ranks of the fully known matrices $W$. 
Therefore, one can provide theoretically sound HOSVD ranks (see Theorem~\ref{thm:hosvd_quality}
in Section~\ref{sec:theory}).
A more detailed comparison with related work is presented in Section~\ref{sec:related_work}. A design comparison of tensor-based PEFT methods is provided in Table~\ref{tab:method_comparison} (Appendix~\ref{app:additional}), along with a visual taxonomy (Figure~\ref{fig:taxonomy}) and parameter scaling analysis (Figure~\ref{fig:param_scaling}).

\section{The CRAFT methodology}
\label{sec:method}

In what follows, we denote tensors by calligraphic letters. We denote elements of matrix $A$ or tensor  $\mathcal{W}$ by $A(i,j)$ and  $\mathcal{W} (i,j,k)$.
Following~\citep{kolda2009tensor}, we introduce the following notation. 
A third-order tensor $\mathcal{W} \in \R^{I_1 \times I_2 \times I_3}$ is a three-dimensional array with elements $w_{i_1 i_2 i_3}$.

\begin{definition}[Mode-$n$ Unfolding]
The mode-$n$ unfolding (matricization) of tensor $\mathcal{W}$, denoted $W_{(n)}$, arranges the mode-$n$ fibers as columns of a matrix. For a third-order tensor:
$W_{(1)} \in \R^{I_1 \times I_2 I_3}$,  $W_{(2)} \in \R^{I_2 \times I_1 I_3}$,  
$W_{(3)} \in \R^{I_3 \times I_1 I_2}$.
\end{definition}

\begin{definition}[Mode-$n$ Product]
The mode-$n$ product of tensor $\mathcal{W} \in \R^{I_1 \times I_2 \times I_3}$ with matrix $U \in \R^{J \times I_n}$, denoted $\mathcal{W} \times_n U$, yields a tensor of size $I_1 \times \cdots \times I_{n-1} \times J \times I_{n+1} \times \cdots \times I_N$. In particular, for $n=2$, one has
$\displaystyle (\mathcal{W} \times_2 U) (i_1,  j, i_3) = \sum_{i_2=1}^{I_2} \mathcal{W}  (i_1, i_2, i_3) \, U (j, i_2).$
\end{definition}

\begin{definition}[Tucker Decomposition]
The Tucker decomposition of $\mathcal{W} \in \R^{I_1 \times I_2 \times I_3}$ with multilinear rank $(r_1, r_2, r_3)$ is:
$\mathcal{W} \approx \mathcal{G} \times_1 U^{(1)} \times_2 U^{(2)} \times_3 U^{(3)}$
where $\mathcal{G} \in \R^{r_1 \times r_2 \times r_3}$ is the core tensor and $U^{(n)} \in \R^{I_n \times r_n}$, $n=1,2,3$,  are factor matrices with orthonormal columns.
\end{definition}

\begin{figure*}[t]
\centering
\begin{tikzpicture}[scale=0.78, transform shape]
    \fill[lightblue, rounded corners=5pt] (-0.5,-0.5) rectangle (17.5,6.5);
    
    \node[font=\large\bfseries] at (8.5,6) {CRAFT Architecture: Cross-layer Tucker Decomposition with Frozen Factors};
    
    \node[font=\small\bfseries] at (1.5,5) {Pre-trained Weights};
    
    \foreach \i in {0, 0.3, 0.6, 0.9} {
        \fill[white, draw=craftblue, thick, rounded corners=2pt] 
            (0.3+\i, 1.5-\i*0.5) rectangle (2.7+\i, 4-\i*0.5);
    }
    \node[font=\small, align=center] at (1.9,2.55) {$W_Q^{(1)},\ldots,$\\$W_Q^{(N_L)}$};
    
    \draw[->, thick, craftblue] (3.5,2.5) -- (4.8,2.5);
    \node[font=\small, above] at (4.15,2.5) {Stack};
    
    \node[font=\small\bfseries] at (6,5) {3D Weight Tensor};
    
    \begin{scope}[shift={(5,1)}]
        \fill[craftorange!30] (0,0) -- (2,0) -- (2.6,0.6) -- (0.6,0.6) -- cycle;
        \fill[craftorange!50] (2,0) -- (2,2.5) -- (2.6,3.1) -- (2.6,0.6) -- cycle;
        \fill[craftorange!70] (0,0) -- (2,0) -- (2,2.5) -- (0,2.5) -- cycle;
        \draw[thick] (0,0) -- (2,0) -- (2,2.5) -- (0,2.5) -- cycle;
        \draw[thick] (2,0) -- (2.6,0.6) -- (2.6,3.1) -- (2,2.5);
        \draw[thick] (0,2.5) -- (0.6,3.1) -- (2.6,3.1);
        \draw[thick, dashed] (0,0) -- (0.6,0.6) -- (2.6,0.6);
        \draw[thick, dashed] (0.6,0.6) -- (0.6,3.1);
        
        \node[font=\small] at (1,-0.3) {$d_{in}$};
        \node[font=\small] at (2.9,1.6) {$d_{out}$};
        \node[font=\small] at (0.3,3.3) {$N_L$};
        \node[font=\small] at (1.3,1.5) {$\mathcal{W}_Q$};
    \end{scope}
    
    \draw[->, thick, craftgreen] (8,2.5) -- (9.3,2.5);
    \node[font=\small, above] at (8.65,2.6) {HOSVD};
    
    \node[font=\small\bfseries] at (12,5) {Frozen Tucker Factors};
    
    \begin{scope}[shift={(9.5,2)}]
        \fill[craftgreen!30] (0,0) -- (0.8,0) -- (1.1,0.3) -- (0.3,0.3) -- cycle;
        \fill[craftgreen!50] (0.8,0) -- (0.8,1) -- (1.1,1.3) -- (1.1,0.3) -- cycle;
        \fill[craftgreen!70] (0,0) -- (0.8,0) -- (0.8,1) -- (0,1) -- cycle;
        \draw[thick] (0,0) -- (0.8,0) -- (0.8,1) -- (0,1) -- cycle;
        \draw[thick] (0.8,0) -- (1.1,0.3) -- (1.1,1.3) -- (0.8,1);
        \draw[thick] (0,1) -- (0.3,1.3) -- (1.1,1.3);
        \node[font=\small] at (0.55,0.5) {$\mathcal{G}$};
    \end{scope}
    
    \fill[craftblue!40, rounded corners=2pt] (10.9,3.2) rectangle (11.6,4.3);
    \draw[thick, craftblue] (10.9,3.2) rectangle (11.6,4.3);
    \node[font=\small] at (11.25,3.75) {$U^{(1)}$};
    
    \fill[craftpurple!40, rounded corners=2pt] (11.9,2.7) rectangle (12.6,4.3);
    \draw[thick, craftpurple] (11.9,2.7) rectangle (12.6,4.3);
    \node[font=\small] at (12.25,3.5) {$U^{(2)}$};
    
    \fill[craftorange!60, rounded corners=2pt] (12.9,2.2) rectangle (13.6,4.3);
    \draw[thick, craftorange] (12.9,2.2) rectangle (13.6,4.3);
    \node[font=\small] at (13.25,3.25) {$U^{(3)}$};
    
    \node[font=\small\bfseries, craftred] at (15.8,5) {Trainable};
    
    \fill[craftred!30, rounded corners=2pt] (14.3,3.35) rectangle (15.1,4.15);
    \draw[thick, craftred] (14.3,3.35) rectangle (15.1,4.15);
    \node[font=\small] at (14.7,3.75) {$J^{(1)}$};
    
    \fill[craftred!30, rounded corners=2pt] (15.25,3.35) rectangle (16.05,4.15);
    \draw[thick, craftred] (15.25,3.35) rectangle (16.05,4.15);
    \node[font=\small] at (15.65,3.75) {$J^{(2)}$};
    
    \fill[craftred!30, rounded corners=2pt] (16.2,3.35) rectangle (17.0,4.15);
    \draw[thick, craftred] (16.2,3.35) rectangle (17.0,4.15);
    \node[font=\small] at (16.6,3.75) {$J^{(3)}$};
    
    \fill[white, rounded corners=3pt, draw=black, thick] (1.0,-0.2) rectangle (16.5,1.2);
    \node[font=\small, align=center] at (8.75,0.5) {
        $\widehat{\mathcal{W}} = \mathcal{W} + \bigl[\mathcal{G} \times_1 (U^{(1)} \colred{J^{(1)}}) \times_2 (U^{(2)} \colred{J^{(2)}}) \times_3 (U^{(3)} \colred{J^{(3)}}) - \mathcal{G} \times_1 U^{(1)} \times_2 U^{(2)} \times_3 U^{(3)}\bigr]$
    };
\end{tikzpicture}
\caption{CRAFT architecture overview. Pre-trained attention weights (Q, V) are stacked across $N_L$ layers into 3D tensors. HOSVD decomposes each tensor into a core tensor $\mathcal{G}$ and factor matrices $U^{(1)}, U^{(2)}, U^{(3)}$. All decomposition factors are \textbf{frozen}. Adaptation occurs only through small square matrices $J^{(1)} \in \mathbb{R}^{r_1 \times r_1}$, $J^{(2)} \in \mathbb{R}^{r_2 \times r_2}$, $J^{(3)} \in \mathbb{R}^{r_3 \times r_3}$ (shown in red), initialized near identity.  This yields a trainable parameter count of \rev{$n_p(r_1^2 + r_2^2 + r_3^2)$, where $n_p$ is the number of adapted projection types ($n_p = 2$ for Q and V)}---\textbf{independent of model dimension $d$ and depth $N_L$} at fixed Tucker ranks.}
\label{fig:architecture}
\end{figure*}

\begin{algorithm}[t]
\caption{CRAFT: Cross-layer Rank Adaptation via Frozen Tucker}
\label{alg:craft}
\small
\begin{algorithmic}[1]
\REQUIRE \rev{Set of adapted projections $\mathcal{A} \subseteq \{Q, K, V, O\}$ with $n_p = |\mathcal{A}|$ (default $\mathcal{A} = \{Q,V\}$);} pre-trained weights $\{W_\alpha^{(\ell)}\}_{\ell=1}^{N_L}$ for $\alpha \in \rev{\mathcal{A}}$, $W_\alpha^{(\ell)} \in \R^{d_{out} \times d_{in}}$; Tucker ranks $(r_1, r_2, r_3)$ \rev{with $r_1 \leq N_L$}; \rev{scaling $s>0$;} \fixnew{dropout rate $p \geq 0$ on $J^{(n)}$;} learning rate~$\eta$; loss $\mathcal{L}$
\ENSURE Adapted weight tensor $\widehat{\mathcal{W}}_\alpha$; frozen factors $U^{(n)}$, core $\mathcal{G}$; trained matrices $J^{(n)}$, $n=1,2,3$
\STATE \textbf{// Stage 1: Tensor Construction}
\FOR{each projection $\alpha \in \rev{\mathcal{A}}$}
    \STATE Stack: $\mathcal{W}_\alpha \leftarrow \text{stack}(W_\alpha^{(1)}, \ldots, W_\alpha^{(N_L)}) \in \R^{N_L \times d_{out} \times d_{in}}$
\ENDFOR
\STATE \textbf{// Stage 2: HOSVD Decomposition (one-time)}
\FOR{each $\alpha \in \rev{\mathcal{A}}$ and $n = 1, 2, 3$}
    \STATE Compute mode-$n$ unfolding $W_{(n)}$; set $U^{(n)}_\alpha \leftarrow$ first $r_n$ left singular vectors of $W_{(n)}$
\ENDFOR
\FOR{each $\alpha \in \rev{\mathcal{A}}$}
    \STATE $\mathcal{G}_\alpha \leftarrow \mathcal{W}_\alpha \times_1 (U^{(1)}_\alpha)^\top \times_2 (U^{(2)}_\alpha)^\top \times_3 (U^{(3)}_\alpha)^\top$;\quad $\mathcal{R}_\alpha \leftarrow \mathcal{G}_\alpha \times_1 U^{(1)}_\alpha \times_2 U^{(2)}_\alpha \times_3 U^{(3)}_\alpha$
    \STATE \textbf{Freeze:} $\mathcal{W}_\alpha,\; \mathcal{G}_\alpha,\; \mathcal{R}_\alpha,\; U^{(1)}_\alpha,\; U^{(2)}_\alpha,\; U^{(3)}_\alpha$ \rev{(in practice, store the single buffer $\mathcal{W}_\alpha - \mathcal{R}_\alpha$)}
\ENDFOR
\STATE \textbf{// Stage 3: Initialize Trainable Adaptation}
\FOR{each $\alpha \in \rev{\mathcal{A}}$, each $n \in \{1,2,3\}$}
    \STATE $J^{(n)}_\alpha \leftarrow I_{r_n} + \epsilon \cdot E^{(n)}$, \quad $E^{(n)}_{ij} \sim \text{i.i.d.}\ \mathcal{N}(0, \sigma^2)$
\ENDFOR
\STATE \textbf{// Stage 4: Training (residual-preserving)}
\WHILE{not converged}
    \FOR{each $\alpha \in \rev{\mathcal{A}}$}
        \STATE $\mathcal{T}_\alpha \leftarrow \mathcal{G}_\alpha \times_1 (U^{(1)}_\alpha \fixnew{\widetilde{J}}^{(1)}_\alpha) \times_2 (U^{(2)}_\alpha \fixnew{\widetilde{J}}^{(2)}_\alpha) \times_3 (U^{(3)}_\alpha \fixnew{\widetilde{J}}^{(3)}_\alpha)$, \quad \fixnew{$\widetilde{J}^{(n)}_\alpha = \text{Dropout}_p(J^{(n)}_\alpha)$}
        \STATE $\widehat{\mathcal{W}}_\alpha \leftarrow \mathcal{W}_\alpha + \rev{s}\,(\mathcal{T}_\alpha - \mathcal{R}_\alpha)$; \quad extract $\widehat{W}_\alpha^{(\ell)} \leftarrow \widehat{\mathcal{W}}_\alpha(\ell,:,:)$
    \ENDFOR
    \STATE Forward pass with $\{\widehat{W}_\alpha^{(\ell)}\}_{\alpha \in \rev{\mathcal{A}}}$ \rev{(projections outside $\mathcal{A}$ frozen)}; compute $\mathcal{L}$; update $J^{(n)}_\alpha \leftarrow J^{(n)}_\alpha - \eta \nabla_{J^{(n)}_\alpha} \mathcal{L}$
\ENDWHILE
\RETURN $\widehat{\mathcal{W}}_\alpha$; $\{U^{(n)}_\alpha, \mathcal{G}_\alpha, J^{(n)}_\alpha\}$, $n=1,2,3$
\end{algorithmic}
\end{algorithm}

Given a transformer with $N_L$ layers, CRAFT proceeds in three stages: tensor construction, frozen HOSVD decomposition, and adaptation via trainable matrices. The complete procedure is described in Algorithm~\ref{alg:craft}.

\textbf{Stage 1: Tensor Construction.} For each projection type $\alpha \in \rev{\mathcal{A}}$, we construct a 3D tensor by stacking the pre-trained attention weight matrices across layers:
\begin{equation*}
\mathcal{W}_\alpha = \text{stack}(W_\alpha^{(1)}, W_\alpha^{(2)}, \ldots, W_\alpha^{(N_L)}) \in \mathbb{R}^{N_L \times d_{out} \times d_{in}}
\end{equation*}
This construction treats the collection of weight matrices as a single multi-way array, enabling decomposition methods that capture correlations simultaneously across layers (mode-1), output dimensions (mode-2), and input dimensions (mode-3).

\textbf{Stage 2: HOSVD Decomposition, Residual Computation, and Freezing.} We compute the Tucker-3 decomposition of each $\mathcal{W}_\alpha$ via HOSVD~\citep{de2000multilinear}. For each mode $n \in \{1,2,3\}$, we compute the mode-$n$ unfolding $(W_\alpha)_{(n)}$, take its truncated SVD to obtain the first $r_n$ left singular vectors $U_\alpha^{(n)}$, and then compute the core tensor $\mathcal{G}_\alpha = \mathcal{W}_\alpha \times_1 (U_\alpha^{(1)})^\top \times_2 (U_\alpha^{(2)})^\top \times_3 (U_\alpha^{(3)})^\top$. \rev{We use standard truncated HOSVD computed from the three mode unfoldings, without a subsequent HOOI refinement step, so that Theorem~\ref{thm:hosvd_quality} applies verbatim. Note that the mode-1 unfolding has only $N_L$ rows, so the mode-1 rank is necessarily capped at $r_1 \leq N_L$.}
Since the Tucker decomposition with ranks $(r_1, r_2, r_3) < (N_L, d_{out}, d_{in})$ is  not exact, the initial reconstruction 
\begin{equation}  \label{eq:R_alpha}
 \mathcal{R}_\alpha = \mathcal{G}_\alpha \times_1 U^{(1)}_\alpha \times_2 U^{(2)}_\alpha \times_3 U^{(3)}_\alpha   
\end{equation}
is \emph{not} equal to $\mathcal{W}_\alpha$ in general. \fixv{We therefore retain the residual $\mathcal{W}_\alpha-\mathcal{R}_\alpha$; together with the adapted Tucker reconstruction, this permits exact recovery of the pre-trained weights when $J_\alpha^{(n)}=I_{r_n}$. With the reported near-identity initialization $J_\alpha^{(n)}=I_{r_n}+10^{-3}E^{(n)}$, the model starts close to, but not exactly at, the pre-trained solution.}
\fixv{The residual buffer $\mathcal{W}_\alpha-\mathcal{R}_\alpha$, the core $\mathcal{G}_\alpha$, and the factor matrices $U^{(1)}_\alpha, U^{(2)}_\alpha, U^{(3)}_\alpha$ are frozen.}

\textbf{Stage 3: Residual-Preserving Adaptation.} We introduce trainable square matrices 
$J^{(n)}_\alpha \in \R^{r_n \times r_n}$, $n=1,2,3$,
initialized near identity:
\begin{equation}
J^{(n)}_\alpha = I_{r_n} + \epsilon \cdot E, \quad E_{ij} \sim \mathcal{N}(0, \sigma^2)
\label{eq:init}
\end{equation}
with \rev{$\epsilon = 10^{-3}$ and $\sigma = 1$; equivalently, every entry of $J^{(n)}_\alpha$ is perturbed by i.i.d.\ Gaussian noise of standard deviation $10^{-3}$}. The adapted weight tensor is computed via the \emph{residual-preserving} formula:
\begin{equation}
\widehat{\mathcal{W}}_\alpha = \mathcal{W}_\alpha + \rev{s}\,\bigl(\mathcal{T}_\alpha - \mathcal{R}_\alpha\bigr),
\label{eq:craft_adaptation}
\end{equation}
where
\begin{equation}
\mathcal{T}_\alpha = \mathcal{G}_\alpha \times_1 (U_\alpha^{(1)} J_\alpha^{(1)}) \times_2 (U_\alpha^{(2)} J_\alpha^{(2)}) \times_3 (U_\alpha^{(3)} J_\alpha^{(3)})
\label{eq:tucker_adapted}
\end{equation}
is the Tucker reconstruction with adapted factors\rev{, and $s > 0$ is an optional scaling factor on the adaptation, set to $s = 1$ in all reported experiments}.
The individual layer weights are extracted as $\widehat{W}_\alpha^{(\ell)} = \widehat{\mathcal{W}}_\alpha(\ell, :, :)$ for $\ell = 1, \ldots, N_L$. 

\begin{remark}[Weight Preservation at Initialization]
\label{rem:preservation}
When $J^{(n)}_\alpha = I_{r_n}$ for all $n$, we have $\mathcal{T}_\alpha = \mathcal{R}_\alpha$, so $\widehat{\mathcal{W}}_\alpha = \mathcal{W}_\alpha$\rev{, for any value of $s$}. The adapted model therefore starts \emph{exactly} at the pre-trained solution, regardless of the Tucker approximation error. This property, analogous to PiSSA's residual design, is critical for stable fine-tuning. \fixnew{This statement concerns the deterministic map \eqref{eq:craft_adaptation}--\eqref{eq:tucker_adapted}} \fixv{at $J^{(n)}_\alpha = I_{r_n}$. In the reported runs the $J^{(n)}_\alpha$ are initialized as $I_{r_n} + 10^{-3}E$ rather than exactly at $I_{r_n}$, so the model starts close to, but not exactly at, the pre-trained solution; the size of this gap is quantified in Theorem~\ref{th:init}, and the effect of the dropout used during training in Remark~\ref{rem:implementation}.}
\end{remark}

\begin{remark}[\rev{Implementation}]
\label{rem:implementation}
\fixv{The implementation stores frozen residual slices and reconstructs each adapted layer weight $\widehat{W}_\alpha^{(\ell)}$ on demand through a weight parametrization of the corresponding linear layer, avoiding materialization of the complete adapted cross-layer tensor.} \fixv{Dropout is applied to $J^{(n)}_\alpha$ before it multiplies $U^{(n)}_\alpha$; all reported LLaMA runs use $p = 0.005$, and dropout is inactive at evaluation time. During training, inverted dropout makes the reconstructed weights stochastic. At $J^{(n)}_\alpha = I_{r_n}$, the expected layer-wise reconstruction equals the corresponding pretrained reconstruction. Exact deterministic weight preservation holds in evaluation mode.} \rev{The HOSVD factors, the core tensor and the matrices $J^{(n)}_\alpha$ are kept in \texttt{float32} for numerical stability of the SVD, and the reconstructed weight is cast back to the model dtype (\texttt{bfloat16} for the LLaMA experiments) in the forward pass.}
\end{remark}

\section{Comparison with related work}
\label{sec:related_work}

{\bf Low-rank adaptation methods. \ }
LoRA~\citep{hu2022lora} decomposes weight updates into low-rank matrices: $\Delta W = BA$. Extensions include AdaLoRA~\citep{zhang2023adalora} with adaptive rank allocation, QLoRA~\citep{dettmers2023qlora} combining quantization with low-rank adaptation, DoRA~\citep{liu2024dora} decomposing weights into magnitude and direction, and GaLore~\citep{zhao2024galore} projecting gradients onto low-rank subspaces. Adapter methods~\citep{houlsby2019parameter, ruckle2021adapterdrop} insert small trainable modules between frozen layers. Despite their success, these approaches generally adapt model components on a per-layer basis and do not explicitly exploit cross-layer tensor structure.

{\bf SVD-based weight decomposition for PEFT.\ }
PiSSA~\citep{meng2024pissa} initializes LoRA adapters with principal singular values and vectors of pre-trained weights while freezing the residual. Related methods include OLoRA~\citep{buyukakyuz2024olora} using QR decomposition and LoRA-XS~\citep{balazy2024loraxs}, which inserts a small trainable $r \times r$ matrix between frozen SVD-derived factors. These methods demonstrate the value of decomposing pre-trained weights but all operate \textit{independently per layer}.

{\bf Tensor-based PEFT methods.\ }
LoTR~\citep{bershatsky2024lotr} applies Tucker-2 decomposition to gradient updates with shared factor matrices across layers. VeRA~\citep{kopiczko2024vera} shares random frozen projection matrices across all layers while training per-layer scaling vectors. LoRETTA~\citep{yang2024loretta} applies tensor-train decomposition to weight updates. SuperLoRA~\citep{chen2024superlora} groups weight updates $\Delta W$ across layers, reshapes them into higher-order tensors, and applies Tucker decomposition. However, like LoTR, SuperLoRA decomposes the \textit{gradient update} tensor. CRAFT is distinct: it decomposes the \textit{pre-trained weights} using full Tucker-3 decomposition via HOSVD, preserves original weights through a residual formulation, and adapts through small trainable transformation matrices applied to each frozen factor.

{\bf Parameter complexity comparison.\ }

\begin{proposition}[Trainable Parameter Count and Scaling Comparison]
\label{prop:params_and_comparison}
CRAFT with Tucker ranks $(r_1, r_2, r_3)$ applied to $n_p$ projection types has exactly
$N_{\text{train}} = n_p \cdot (r_1^2 + r_2^2 + r_3^2)$
Tucker adaptation parameters (excluding the task-specific classifier head), independent of model dimension $d$ and depth $N_L$ for fixed ranks. For a model with $N_L$ layers and dimension $d$ (where $d_{out} \asymp d_{in} \asymp d$), the trainable parameter counts of competing methods are:
\textbf{(1)} Full fine-tuning: $O(N_L \cdot d^2)$;
\textbf{(2)} LoRA/PiSSA (rank $r$): $O(N_L \cdot r \cdot d)$;
\textbf{(3)} LoTR: $O(N_L \cdot r^2 + r \cdot d)$;
\textbf{(4)} CRAFT: $O(r_1^2 + r_2^2 + r_3^2)$.
CRAFT is the only method among those compared whose complexity is independent of both $N_L$ and $d$ at fixed ranks.
\end{proposition}

\begin{remark}
In the GLUE experiments, task-specific classifier parameters are also trained. The 40K--41K figures reported for GLUE refer only to the CRAFT-specific Tucker adaptation parameters and exclude the task-specific classifier parameters. \rev{In the LLaMA commonsense experiments no classification head is introduced and the language-modeling head remains frozen, so the matrices $J^{(n)}_\alpha$ are the \emph{only} trainable parameters.}
\end{remark}

\section{Theoretical analysis}
\label{sec:theory}

We provide theoretical foundations for CRAFT, including approximation guarantees, expressiveness characterization, and optimization properties. \fixw{Proofs of Proposition~\ref{prop:params_and_comparison}, Theorem~\ref{th:init} and Theorem~\ref{thm:gradient} are deferred to Appendix~\ref{app:proofs}; Theorem~\ref{thm:hosvd_quality} is quoted from~\citet{de2000multilinear} and Theorem~\ref{thm:subspace} is immediate from the construction \eqref{eq:craft_adaptation}--\eqref{eq:new_tucker}.}

\medskip


\begin{theorem}[{\bf HOSVD Quasi-Optimality} ~\citep{de2000multilinear}]
\label{thm:hosvd_quality}
Let $\mathcal{W} \in \R^{I_1 \times I_2 \times I_3}$ with mode-$n$ unfolding singular values 
$\sigma_1^{(n)} \geq \sigma_2^{(n)} \geq \cdots \geq \sigma_{I_n}^{(n)}$.  
Let \fixw{$\widetilde{\mathcal{W}}$} be its rank-$(r_1, r_2, r_3)$ HOSVD approximation and 
$\mathcal{W}^*$ be the best rank-$(r_1, r_2, r_3)$ Tucker approximation. Then 
$$
\fnorm{\mathcal{W} - \fixw{\widetilde{\mathcal{W}}}} \leq \sqrt{3} \cdot \fnorm{\mathcal{W} - \mathcal{W}^*}, \quad
\fnorm{\mathcal{W} - \fixw{\widetilde{\mathcal{W}}}}^2 \leq \sum_{n=1}^{3} \sum_{j=r_n+1}^{I_n} \left(\sigma_j^{(n)}\right)^2.
$$
Moreover, if the singular values of the mode-$n$ unfoldings decay as $\sigma_j^{(n)} = O(j^{-\beta})$ for some $\beta > 1/2$, 
then \fixv{the squared tail satisfies $\sum_{j > r_n} (\sigma_j^{(n)})^2 = O(r_n^{1-2\beta})$, so that ranks $r_n = O\bigl(\epsilon^{-1/(\beta - 1/2)}\bigr)$} suffice for $\epsilon$-approximation in the Frobenius norm.
\end{theorem}


%
We consider weight adaptation of the forms \eqref{eq:craft_adaptation}
and \eqref{eq:tucker_adapted}. Therefore, one can rewrite $\mathcal{T}_\alpha$ as  
\begin{equation}  \label{eq:new_tucker}
 \mathcal{T}_\alpha = \widehat{\mathcal{G}}_\alpha \times_1  U_\alpha^{(1)}  \times_2  U_\alpha^{(2)}   \times_3  U_\alpha^{(3)}, 
 \quad
\widehat{\mathcal{G}}_\alpha =  \mathcal{G}_\alpha \times_1  J_\alpha^{(1)}  \times_2   J_\alpha^{(2)}  \times_3   J_\alpha^{(3)}. 
\end{equation}
Consider a set $\mathcal{S}$ of orbits of tensor $\mathcal{G}_\alpha$:
$$
\mathcal{S} (\mathcal{G}_\alpha) = \left\{ \mathcal{G}_\alpha \times_1  J^{(1)}  \times_2   J^{(2)}  \times_3   J^{(3)}: J^{(n)} \in GL(r_n) \right\}
$$ 
where $GL(r_n)$ denotes the group of invertible $r_n \times r_n$ matrices, $n=1,2,3$. Then, the collection of tensors $\mathcal{T}_\alpha$
in \eqref{eq:new_tucker} forms a smooth manifold of dimension \rev{at most} $r_1^2 + r_2^2 + r_3^2$ embedded in $\R^{N_L \times d_{out} \times d_{in}}$\rev{; the bound is not attained exactly because the triple $(J^{(1)}, J^{(2)}, J^{(3)})$ and $(c_1 J^{(1)}, c_2 J^{(2)}, c_3 J^{(3)})$ with $c_1 c_2 c_3 = 1$ produce the same tensor, so the stabilizer of the action is at least two-dimensional}.

 
\begin{theorem}[{\bf Orbit-Constrained Expressiveness}]
\label{thm:subspace}
Let $\mathcal{W}$ have HOSVD decomposition with factors $U^{(n)}$, $n=1,2,3$. Then CRAFT can adapt to any tensor $\mathcal{T}_\alpha$
of the form \eqref{eq:new_tucker} with $\widehat{\mathcal{G}}_\alpha \in \mathcal{S} (\mathcal{G}_\alpha)$.
Moreover, if $J^{(n)} = I_{r_n}$ for $n=1,2,3$, then CRAFT exactly recovers the original weights:
$\widehat{\mathcal{W}}_\alpha = \mathcal{W}_\alpha$.
\end{theorem}


\begin{theorem}[{\bf Near-Identity Initialization Effect}]
\label{th:init}
Let $\mathcal{R}_\alpha$ be of the form \eqref{eq:R_alpha} and 
$\mathcal{T}_\alpha^{\text{ini}}  = \mathcal{G}^{\text{ini}}_\alpha \times_1  U_\alpha^{(1)}  
\times_2  U_\alpha^{(2)}   \times_3  U_\alpha^{(3)} $.
Consider initialization with $J^{(n)} = I_{r_n} + \epsilon E^{(n)}$ where $E^{(n)}_{i,j} \sim \mathcal{N}(0, \sigma^2)$\rev{, $\sigma = 1$}, $n=1,2,3$,
\begin{equation}  \label{eq:G_alpha_ini}
\mathcal{G}_\alpha^{\text{ini}} = \mathcal{G}_\alpha \times_1 J^{(1)} \times_2 J^{(2)} \times_3 J^{(3)},
\end{equation} 
and let $r = \max(r_1, r_2, r_3)$. 
Then, for any $\delta \in (0,1)$, with probability at least $1 - 8 \delta$,
\begin{equation}  \label{eq:R_ini_err}
\fnorm{\mathcal{T}_\alpha^{\text{ini}} - \mathcal{R}_\alpha} \leq C(r,\epsilon, \delta)\, \fnorm{\mathcal{R}_\alpha},
\end{equation}
where $C(r,\epsilon, \delta) = O\bigl(\epsilon(\sqrt{r} + \sqrt{\log \delta^{-1}})\bigr)$ for small $\epsilon$. Precise constants are given in Appendix~\ref{app:proofs}. \rev{With $\epsilon = 10^{-3}$ and the ranks used in our experiments, the right-hand side of \eqref{eq:R_ini_err} stays below $0.3\,\fnorm{\mathcal{R}_\alpha}$ (a worked numerical example is given in Appendix~\ref{app:proofs}), so the model starts in a small neighbourhood of the pre-trained solution.}
\end{theorem}

\subsection{Optimization properties}

\begin{theorem}[Gradient Flow at Initialization]
\label{thm:gradient}
At initialization $(J^{(1)}, J^{(2)}, J^{(3)})$,
the gradient of loss $\mathcal{L}$ with respect to $J^{(1)}$ is:
$ \displaystyle
\frac{\partial \mathcal{L}}{\partial J^{(1)}} = U^{(1)\top} \frac{\partial \mathcal{L}}{\partial U^{(1)}_{\text{eff}}},
$
where $U^{(1)}_{\text{eff}} = U^{(1)} J^{(1)}$ is the effective factor matrix. Since $U^{(1)}$ has orthonormal columns, $U^{(1)\top}$ \rev{does not amplify gradient norms}:
\begin{equation}
\fnorm{\frac{\partial \mathcal{L}}{\partial J^{(1)}}} = \fnorm{U^{(1)\top} \frac{\partial \mathcal{L}}{\partial U^{(1)}_{\text{eff}}}} \leq \fnorm{\frac{\partial \mathcal{L}}{\partial U^{(1)}_{\text{eff}}}}
\end{equation}
with equality when the gradient lies in the column space of $U^{(1)}$.
\end{theorem}

\section{Experimental results}
\label{sec:experiments}

\subsection{Experimental setup}

We evaluate CRAFT on two sets of benchmarks: (i) the GLUE benchmark~\citep{wang2018glue} using RoBERTa-base (125M parameters, 12 layers) and RoBERTa-large (355M parameters, 24 layers)~\citep{liu2019roberta}, and (ii) eight commonsense reasoning datasets using LLaMA2-7B and LLaMA3-8B. We follow the experimental protocol from~\citet{bershatsky2024lotr} for fair comparison on GLUE. 

For RoBERTa experiments, CRAFT is applied to Q and V projections with \fixw{mode-2 and mode-3 ranks $r_2 = r_3 = 100$ and mode-1 rank $r_1 = N_L$, which is the largest admissible value: $(r_1,r_2,r_3) = (24,100,100)$ for RoBERTa-large ($N_L = 24$), giving $2\times(24^2+100^2+100^2) = 41{,}152 \approx 41\text{K}$ Tucker adaptation parameters, and $(12,100,100)$ for RoBERTa-base ($N_L = 12$), giving $2\times(12^2+100^2+100^2) = 40{,}288 \approx 40\text{K}$. Both are reported as $0.04$M in Table~\ref{tab:glue_results}.} For LLaMA experiments \rev{(both models have $N_L = 32$ layers)}, we use Tucker ranks $(r_1, r_2, r_3) = (32, 128, 128)$, yielding $N_{\text{train}} = 2 \times (32^2 + 128^2 + 128^2) = 67{,}584 \approx 68\text{K}$ Tucker adaptation parameters\fixnew{, reported as 0.068M throughout}. We also report results with smaller ranks \fixnew{$(32, 64, 64)$} corresponding to \fixnew{$2 \times (32^2 + 64^2 + 64^2) = 18{,}432 \approx 18\text{K}$} parameters\rev{, with larger ranks $(32, 250, 250)$ corresponding to $252{,}048 \approx 0.25\text{M}$ parameters, and with the CRAFT$^\star$ configuration, which adapts all four projections $\{Q,K,V,O\}$ with ranks $(32, 360, 360)$ and therefore uses $4 \times (32^2 + 360^2 + 360^2) = 1{,}040{,}896 \approx 1.0\text{M}$ parameters}. \fixnew{Note that $r_1 = 32 = N_L$ for both LLaMA backbones, so the mode-1 factor is a full orthogonal matrix and mode~1 is not truncated in these runs; the cross-layer structure is exploited through the core tensor and the trainable $J^{(1)}$ rather than through mode-1 rank reduction.}

\rev{\textbf{Commonsense training protocol.} We follow the LLM-Adapters protocol: training on the Commonsense-170K instruction dataset with the Alpaca-style prompt template, cross-entropy loss computed on response tokens only, and generative per-task evaluation with beam search (4 beams, at most 32 new tokens) followed by regular-expression answer extraction. We train for 3 epochs with AdamW, learning rate $3 \times 10^{-4}$, per-device batch size 4 with gradient accumulation 4 (effective batch size 16), maximum sequence length 256, warmup ratio 0.06, no weight decay,} \fixnew{dropout $p = 0.005$ on the matrices $J^{(n)}_\alpha$, adaptation scaling $s = 1$,} \rev{bfloat16 compute, and seed} \fixnew{25}\rev{. We report, for each model, the epoch achieving the highest accuracy averaged over the eight tasks.} \fixw{In all six Q,V runs, for which complete logs are released, this was the final (third) epoch, so those numbers coincide with a fixed three-epoch protocol.}

\subsection{Commonsense reasoning results}

\begin{table*}[t]
\centering
\small
\setlength{\tabcolsep}{2.2pt}
\begin{tabular}{@{}l r c c c c c c c c c@{}}
\toprule
\textbf{Method} & \textbf{\# Params} & \textbf{BoolQ} & \textbf{PIQA} & \textbf{SIQA} & \textbf{Hella} & \textbf{Wino} & \textbf{ARC-e} & \textbf{ARC-c} & \textbf{OBQA} & \textbf{Avg.} \\
\midrule
\multicolumn{11}{@{}l}{\textbf{LLaMA2-7B}} \\
LoRA$^\ddagger$          & 56M    & 69.8 & 79.9 & 79.5 & 83.6 & 82.6 & 79.8 & 64.7 & 81.0 & 77.6 \\
PiSSA$^\S$               & 56M    & 67.6 & 78.1 & 78.4 & 76.6 & 78.0 & 75.8 & 60.2 & 75.6 & 73.8 \\
MiLoRA$^\S$              & 56M    & 67.6 & 83.8 & 80.1 & 88.2 & 82.0 & 82.8 & 68.8 & 80.6 & 79.2 \\
DoRA$^\|$                & 56M    & 71.8 & 83.7 & 76.0 & 89.1 & 82.6 & 83.7 & 68.2 & 82.4 & 79.7 \\
LoRA-XS$^\ddagger$       & 3.67M  & 72.1 & 83.7 & 80.5 & 86.0 & 83.9 & 86.0 & 71.2 & 81.0 & 80.5 \\
LoRA-XS$^\ddagger$       & 0.92M  & 70.4 & 83.0 & 80.2 & 82.3 & 82.7 & 84.4 & 68.6 & 80.8 & 79.0 \\
LoRA-XS$^\ddagger$       & 0.23M  & 67.2 & 81.8 & 78.1 & 75.4 & 80.8 & 81.2 & 65.9 & 74.6 & 75.6 \\
\cmidrule{1-11}
\cellcolor{blue!8} CRAFT & \cellcolor{blue!8} \textbf{0.018M} & \cellcolor{blue!8} 66.0 & \cellcolor{blue!8} 78.2 & \cellcolor{blue!8} 73.0 & \cellcolor{blue!8} 66.0 & \cellcolor{blue!8} 73.2 & \cellcolor{blue!8} 80.3 & \cellcolor{blue!8} 60.1 & \cellcolor{blue!8} 64.6 & \cellcolor{blue!8} 70.2 \\
\cellcolor{blue!8} CRAFT & \cellcolor{blue!8} \textbf{0.068M} & \cellcolor{blue!8} 68.3 & \cellcolor{blue!8} 81.1 & \cellcolor{blue!8} 77.3 & \cellcolor{blue!8} 80.1 & \cellcolor{blue!8} 77.5 & \cellcolor{blue!8} 83.5 & \cellcolor{blue!8} 63.9 & \cellcolor{blue!8} 73.6 & \cellcolor{blue!8} 75.7 \\
\cellcolor{blue!8} CRAFT & \cellcolor{blue!8} \textbf{0.252M} & \cellcolor{blue!8} 70.1 & \cellcolor{blue!8} 82.2 & \cellcolor{blue!8} 79.9 & \cellcolor{blue!8} 85.5 & \cellcolor{blue!8} 82.0 & \cellcolor{blue!8} 85.5 & \cellcolor{blue!8} 70.0 & \cellcolor{blue!8} 82.6 & \cellcolor{blue!8} 79.7 \\
\cmidrule{1-11}
\cellcolor{green!10} CRAFT$^\star$ & \cellcolor{green!10} \textbf{1.04M} & \cellcolor{green!10} 71.8 & \cellcolor{green!10} 83.1 & \cellcolor{green!10} \textbf{82.5} & \cellcolor{green!10} 85.6 & \cellcolor{green!10} \textbf{84.0} & \cellcolor{green!10} 85.6 & \cellcolor{green!10} 69.5 & \cellcolor{green!10} 82.2 & \cellcolor{green!10} 80.5 \\
\midrule
\multicolumn{11}{@{}l}{\textbf{LLaMA3-8B}} \\
LoRA$^\ddagger$          & 57M    & 70.8 & 85.2 & 79.9 & 91.7 & 84.3 & 84.2 & 71.2 & 79.0 & 80.8 \\
PiSSA$^\S$               & 57M    & 67.1 & 81.1 & 77.2 & 83.6 & 78.9 & 77.7 & 63.2 & 74.6 & 75.4 \\
MiLoRA$^\S$              & 57M    & 68.8 & 86.7 & 77.2 & 92.9 & 85.6 & 86.8 & 75.5 & 81.8 & 81.9 \\
DoRA$^\|$                & 57M    & 74.6 & 89.3 & 79.9 & 95.5 & 85.6 & 90.5 & 80.4 & 85.8 & 85.2 \\
LoRA-XS$^\ddagger$       & 3.67M  & 73.2 & 88.3 & 82.5 & 94.1 & 87.1 & 90.9 & 80.5 & 85.6 & 85.3 \\
LoRA-XS$^\ddagger$       & 0.92M  & 72.8 & 87.4 & 81.5 & 94.2 & 87.4 & 91.3 & 79.2 & 85.0 & 84.8 \\
LoRA-XS$^\ddagger$       & 0.23M  & 66.6 & 85.8 & 79.4 & 90.1 & 85.2 & 87.0 & 76.5 & 81.8 & 81.6 \\
\cmidrule{1-11}
\cellcolor{blue!8} CRAFT & \cellcolor{blue!8} \textbf{0.018M} & \cellcolor{blue!8} 68.5 & \cellcolor{blue!8} 85.7 & \cellcolor{blue!8} 75.8 & \cellcolor{blue!8} 85.7 & \cellcolor{blue!8} 78.3 & \cellcolor{blue!8} 88.9 & \cellcolor{blue!8} 77.9 & \cellcolor{blue!8} 78.2 & \cellcolor{blue!8} 79.9 \\
\cellcolor{blue!8} CRAFT & \cellcolor{blue!8} \textbf{0.068M} & \cellcolor{blue!8} 71.0 & \cellcolor{blue!8} 87.1 & \cellcolor{blue!8} 78.9 & \cellcolor{blue!8} 91.0 & \cellcolor{blue!8} 83.2 & \cellcolor{blue!8} 90.3 & \cellcolor{blue!8} 79.2 & \cellcolor{blue!8} 82.4 & \cellcolor{blue!8} 82.9 \\
\cellcolor{blue!8} CRAFT & \cellcolor{blue!8} \textbf{0.252M} & \cellcolor{blue!8} 73.0 & \cellcolor{blue!8} 87.6 & \cellcolor{blue!8} 80.5 & \cellcolor{blue!8} 93.4 & \cellcolor{blue!8} 87.2 & \cellcolor{blue!8} 91.8 & \cellcolor{blue!8} 80.4 & \cellcolor{blue!8} 85.4 & \cellcolor{blue!8} 84.9 \\
\cmidrule{1-11}
\cellcolor{green!10} CRAFT$^\star$ & \cellcolor{green!10} \textbf{1.04M} & \cellcolor{green!10} 73.7 & \cellcolor{green!10} 88.2 & \cellcolor{green!10} 82.2 & \cellcolor{green!10} 94.0 & \cellcolor{green!10} 86.5 & \cellcolor{green!10} \textbf{92.4} & \cellcolor{green!10} \textbf{80.6} & \cellcolor{green!10} \textbf{86.8} & \cellcolor{green!10} \textbf{85.6} \\
\bottomrule
\end{tabular}
\vspace{2pt}
\\
{\small All baseline results are taken from the original papers: $^\ddagger$LoRA and LoRA-XS from~\citet{balazy2024loraxs}; $^\|$DoRA from~\citet{liu2024dora}; $^\S$PiSSA and MiLoRA from~\citet{wang2024milora}. $^\star$Q,K,V,O adapted, ranks $(32,360,360)$; all other rows adapt Q,V only.}
\caption{Commonsense reasoning accuracy for fine-tuned LLaMA2-7B and LLaMA3-8B across eight datasets. Baseline numbers for LoRA, PiSSA, MiLoRA, DoRA, and LoRA-XS are reported directly from their original papers. The rank for LoRA is 32; ranks for LoRA-XS are 128, 64, and 32. DoRA uses the same rank and adapted modules as LoRA. CRAFT (ours) uses 0.018M--0.252M Tucker parameters on Q and V, independent of model dimension and depth\fixnew{, at Tucker ranks $(32,64,64)$, $(32,128,128)$ and $(32,250,250)$ respectively}. The highlighted CRAFT$^\star$ rows additionally adapt K and O, using $\sim$1M parameters ($\sim$2\% of the 56M/57M baselines). \fixw{Bold in the accuracy columns marks CRAFT$^\star$ entries that attain the best value in their column within the model block; column maxima attained by a baseline are left unmarked. Bold in the \# Params column simply identifies our rows.}}
\label{tab:commonsense_results}
\end{table*}

Table~\ref{tab:commonsense_results} presents results on eight commonsense reasoning benchmarks. On LLaMA2-7B, CRAFT achieves \rev{75.7} average accuracy with only \rev{0.068M} Tucker adaptation parameters  significantly fewer than LoRA's 56M parameters. These results indicate that the cross-layer Tucker structure is particularly effective for tasks that benefit from consistent cross-layer representations.



\subsection{GLUE benchmark results}

\begin{table*}[t]
\centering
\small
\begin{tabular}{l@{\hspace{4pt}}l@{\hspace{6pt}}r@{\hspace{6pt}}c@{\hspace{5pt}}c@{\hspace{5pt}}c@{\hspace{5pt}}c@{\hspace{5pt}}c@{\hspace{5pt}}c@{\hspace{5pt}}c@{\hspace{5pt}}c@{\hspace{5pt}}c}
\toprule
\textbf{Model} & \textbf{Method} & \textbf{\# Params} & \textbf{MNLI} & \textbf{SST-2} & \textbf{MRPC} & \textbf{CoLA} & \textbf{QNLI} & \textbf{QQP} & \textbf{RTE} & \textbf{STS-B} & \textbf{Avg.} \\
\midrule
\multirow{7}{*}{\rotatebox{90}{RoBERTa\textsubscript{base}}}
& FT$^*$              & 125.0M & 87.6 & 94.8 & 90.2 & 63.6 & 92.8 & 91.9 & 78.7 & 91.2 & 86.4 \\
& BitFit$^*$ & 0.1M   & 84.7 & 93.7 & 92.7 & 62.0 & 91.8 & 84.0 & 81.5 & 90.8 & 85.2 \\
& Adpt\textsuperscript{D}$^*$ & 0.3M & 87.1 & 94.2 & 88.5 & 60.8 & 93.1 & 90.2 & 71.5 & 89.7 & 84.4 \\
& Adpt\textsuperscript{D}$^*$ & 0.9M & 87.3 & 94.7 & 88.4 & 62.6 & 93.0 & 90.6 & 75.9 & 90.3 & 85.4 \\
& LoRA$^*$            & 0.3M   & 87.5 & 95.1 & 89.7 & 63.4 & 93.3 & 90.8 & 86.6 & 91.5 & 87.2 \\
& VeRA$^\S$           & 0.04M  & -- & 94.6 & 89.5 & 65.6 & 91.8 & -- & 78.7 & 90.7 & 85.2$^\diamondsuit$ \\
\cmidrule{2-12}
& \cellcolor{blue!8} CRAFT (ours) & \cellcolor{blue!8} \textbf{0.04M} & \cellcolor{blue!8} 85.7 & \cellcolor{blue!8} 95.1 & \cellcolor{blue!8} 89.2 & \cellcolor{blue!8} 59.0 & \cellcolor{blue!8} 92.0 & \cellcolor{blue!8} 89.0 & \cellcolor{blue!8} 75.8 & \cellcolor{blue!8} 90.4 & \cellcolor{blue!8} 84.5  \\
\midrule
\multirow{9}{*}{\rotatebox{90}{RoBERTa\textsubscript{large}}}
& FT$^*$              & 355.0M & 90.2 & 96.4 & 90.9 & 68.0 & 94.7 & 92.2 & 86.6 & 92.4 & 88.9 \\
& LoRA$^*$            & 0.8M   & 90.6 & 96.2 & 90.9 & 68.2 & 94.9 & 91.6 & 87.4 & 92.6 & 89.0 \\
& Adpt\textsuperscript{P}$^\dagger$ & 3.0M & 90.2 & 96.1 & 90.2 & 68.3 & 94.8 & 91.9 & 83.8 & 92.1 & 88.4 \\
& Adpt\textsuperscript{P}$^\dagger$ & 0.8M & 90.5 & 96.6 & 89.7 & 67.8 & 94.8 & 91.7 & 80.1 & 91.9 & 87.9 \\
& Adpt\textsuperscript{H}$^\dagger$ & 6.0M & 89.9 & 96.2 & 88.7 & 66.5 & 94.7 & 92.1 & 83.4 & 91.0 & 87.8 \\
& Adpt\textsuperscript{H}$^\dagger$ & 0.8M & 90.3 & 96.3 & 87.7 & 66.3 & 94.7 & 91.5 & 72.9 & 91.5 & 86.4 \\
& LoRA$^\dagger$      & 0.8M   & 90.6 & 96.2 & 90.2 & 68.2 & 94.8 & 91.6 & 85.2 & 92.3 & 88.6 \\
& VeRA$^\S$           & 0.06M  & -- & 96.1 & 90.9 & 68.0 & 94.4 & -- & 85.9 & 91.7 & 87.8$^\diamondsuit$ \\
\cmidrule{2-12}
& \cellcolor{blue!8} CRAFT (ours) & \cellcolor{blue!8} \textbf{0.04M} & \cellcolor{blue!8} 90.2 & \cellcolor{blue!8} 96.2 & \cellcolor{blue!8} 90.2 & \cellcolor{blue!8} 67.7 & \cellcolor{blue!8} 94.7 & \cellcolor{blue!8} 89.7 & \cellcolor{blue!8} 84.2 & \cellcolor{blue!8} 91.4 & \cellcolor{blue!8} 88.0  \\
\bottomrule
\end{tabular}
\vspace{2pt}
\\
{\small $^\diamondsuit$Average computed over the 6 tasks reported by VeRA (excluding MNLI and QQP). \fixw{The Avg.\ column for all other rows, including CRAFT, is over all eight tasks; the corresponding six-task averages for CRAFT are 83.6 (base) and 87.4 (large), quoted in the text.}}
\caption{RoBERTa with different adaptation methods on the GLUE benchmark. Higher is better. $^*$~indicates numbers taken from~\citet{hu2022lora}, \rev{including}~\citep{zaken2022bitfit}. $^\dagger$~indicates numbers from~\citet{hu2022lora}, using a restricted setup similar to~\citet{houlsby2019parameter}. $^\S$~indicates numbers from~\citet{kopiczko2024vera}, as reported in~\citet{li2024vblora}. VeRA reports on 6 GLUE tasks (excluding MNLI and QQP); we report ``--'' for missing tasks and compute the average over the 6 reported tasks for VeRA. CRAFT rows (highlighted) show our results with only \textbf{0.04M} Tucker adaptation parameters (excluding the task-specific classifier head).}
\label{tab:glue_results}
\end{table*}

Table~\ref{tab:glue_results} shows that CRAFT achieves competitive performance with significantly fewer Tucker adaptation parameters:

\textbf{RoBERTa-large:} CRAFT achieves an 88.0 average score, \rev{coming within 0.4 points of} the 3M-parameter Adpt\textsuperscript{P} adapter \rev{(88.4)} while using ${\sim}75{\times}$ fewer Tucker adaptation parameters. Compared to LoRA (0.8M parameters), CRAFT uses ${\sim}20{\times}$ fewer parameters with only a 1.0-point lower average. \rev{On the six tasks reported by VeRA~\citep{kopiczko2024vera}, CRAFT averages 87.4 against VeRA's 87.8, i.e.\ 0.4 points lower with $1.5\times$ fewer parameters; we stress that CRAFT's 88.0 figure is an eight-task average and is therefore not directly comparable to VeRA's six-task average.}

\textbf{RoBERTa-base:} CRAFT achieves an 84.5 average with 0.04M Tucker adaptation parameters, compared to 87.2 for LoRA with 0.3M parameters, using ${\sim}7{\times}$ fewer parameters. \rev{On the six-task VeRA subset CRAFT averages 83.6 against VeRA's 85.2, so on the smaller backbone the constrained adaptation space is clearly a limitation.}

\textbf{Efficiency--accuracy trade-off:} CRAFT occupies an extreme point on the efficiency--accuracy Pareto frontier. The method is particularly effective on larger models: on RoBERTa-large, the gap to the best baseline narrows to 1.0 point, and on LLaMA3-8B, CRAFT surpasses LoRA despite using \fixnew{$840\times$} fewer parameters (Table~\ref{tab:commonsense_results}).

\textbf{Comparison with LoTR.} A direct comparison with LoTR~\citep{bershatsky2024lotr} in a single table is not straightforward due to different experimental protocols\rev{, and the two averages below are computed over different task subsets}. Nevertheless, the results reported in~\citet{bershatsky2024lotr} are informative: on RoBERTa-large, LoTR with 328K parameters achieves an 83.0 average, while CRAFT achieves 88.0 with only 41K parameters. \fixw{These figures are quoted as published and are not a controlled comparison: the evaluated task subset, the training protocol and the hyperparameter search all differ between the two papers.} These published results motivate a future controlled comparison between decomposing pre-trained weights and decomposing parameter updates.

\section{Advantages, limitations, and discussion}
\label{sec:discussion}

\textbf{Extreme parameter efficiency.} CRAFT achieves accuracy comparable to methods with \rev{7--}\fixnew{840}\rev{$\times$} more trainable parameters. The Tucker adaptation parameter count $n_p(r_1^2 + r_2^2 + r_3^2)$ depends only on the chosen Tucker ranks and the number of adapted projection types. We emphasize that this independence holds for \emph{fixed} Tucker ranks; whether the same ranks suffice as models grow substantially remains an open empirical question\fixnew{, and the mode-1 rank is in any case constrained by $r_1 \leq N_L$}. Our LLaMA experiments provide initial evidence: increasing ranks from \fixnew{$(32,64,64)$} to $(32,128,128)$ on LLaMA3-8B improved average accuracy from \rev{79.9 to 82.9}, and a further increase to $(32,250,250)$ reaches \rev{84.9}. \fixw{We are not aware of another PEFT method reaching comparable accuracy at this parameter budget on models of this scale.}

\textbf{Limitations.} Further evaluation on generation tasks (e.g., instruction following, summarization) and careful \rev{hyperparameter} selection would strengthen the empirical contribution. The HOSVD pre-computation adds a one-time cost \fixw{dominated by the mode-2 and mode-3 unfolding SVDs, which act on $d \times (N_L d)$ matrices and therefore cost $O(N_L d^3)$}. On RoBERTa-base, CRAFT shows a 2.7-point gap compared to LoRA, suggesting the constrained adaptation space may limit performance on smaller models. Results are reported for a single seed\rev{, and for the commonsense benchmarks the reported epoch is the one with the best average accuracy across the eight evaluation sets}\fixw{; in each of the six released Q,V runs this coincided with the final epoch, so those numbers are not inflated by epoch selection}. \rev{For LLaMA2-7B at the smallest budgets, ARC-c is the weakest task relative to the baselines; increasing the Tucker adaptation budget from 0.068M to 0.252M improves ARC-c from 63.9 to 70.0 and ARC-e from 83.5 to 85.5.} 

\section{Conclusion}

We introduced CRAFT, a parameter-efficient fine-tuning method that bridges pre-trained weight decomposition with cross-layer tensor structure. The key contributions of this work are as follows:

\textbf{1.\ Novel framework.} CRAFT combines frozen HOSVD factors with trainable adaptation matrices and a residual-preserving formulation that \fixv{recovers the pre-trained weights exactly when the adaptation matrices equal the identity, so that training begins in a small neighbourhood of the pre-trained solution}.

\textbf{2.\ Extreme efficiency.} With only \rev{40K--68K} Tucker adaptation parameters at fixed ranks---independent of model dimension and depth---CRAFT is, to the best of our knowledge, the only PEFT method among those compared with this property. On LLaMA3-8B, CRAFT with \rev{0.068M} parameters achieves \rev{an 82.9} average on commonsense reasoning, surpassing LoRA (80.8) which uses \fixnew{$840\times$} more parameters.

\section*{Acknowledgments}

\rev{The first two authors gratefully acknowledge partial support by National Science Foundation (NSF) grant DMS-2014928.}

\section*{Ethics statement}

This work proposes a parameter-efficient fine-tuning method. We do not foresee specific ethical concerns beyond those generally associated with fine-tuning large language models.

\section*{Use of large language models}

The authors used large language model tools for assistance with text editing, citation formatting, and coding assistance during the preparation of this manuscript. All LLM-generated outputs were reviewed and verified by the authors, who take full responsibility for the content of this paper.

\bibliographystyle{colm2026_conference}

\appendix

\section{Additional material}
\label{app:additional}

\subsection{Choice of adapted projections}
\label{sec:empirical}

Following the original LoRA framework~\citep{hu2022lora}, CRAFT adapts the Q and V projection matrices while keeping K and O frozen\rev{ in its main configuration}. This choice is motivated by three observations:

\textbf{Gauge invariance of $QK^\top$.}
The attention score matrix is computed as $\text{softmax}(X W_Q W_K^\top X^\top / \sqrt{d_k})$. For any invertible matrix $M \in GL(d_k)$, replacing $W_Q \to W_Q M$ and $W_K \to W_K M^{-\top}$ leaves the product $W_Q W_K^\top$ unchanged. This gauge invariance means that, under unconstrained perturbations, simultaneously adapting both Q and K introduces a redundant degree of freedom, motivating restriction to only one of the two.

\textbf{Independence of the value pathway.}
The attention output for a single head is $\text{Attn}(X) = \text{softmax}(X W_Q W_K^\top X^\top / \sqrt{d_k})\, X W_V$. Since $W_V$ does not enter the softmax argument, it controls an independent subspace: perturbations to $W_V$ change \emph{what} information is read out, while perturbations to $W_Q$ change \emph{where} the model attends. Adapting both Q and V therefore covers complementary degrees of freedom.

\textbf{Empirical PCA analysis.} 
PCA analysis of vectorized attention weights across ViT-B/16, RoBERTa-base, and GPT-2 (Figure~\ref{fig:pca_analysis}) shows that Q weights consistently exhibit greater dispersion across layers compared to K and V, which tend to cluster near the projected mean. Specifically, for each layer $\ell$, let $w_{\alpha,i}^{(\ell)} \in \R^{d_{\mathrm{in}}}$ denote the $i$-th row of projection matrix $W_\alpha^{(\ell)}$. We pool all rows from Q, K, V, compute the sample covariance, and project onto its leading eigenvectors $P_k$. The \emph{dispersion} of projection type $\alpha$ is
\begin{equation}
\sigma_\alpha^{(\ell)}(k) = \Bigl( \frac{1}{d_{\mathrm{out}}} \sum_{i=1}^{d_{\mathrm{out}}} \bigl\| P_k\bigl(w_{\alpha,i}^{(\ell)} - \bar\mu^{(\ell)}\bigr)\bigr\|^2 \Bigr)^{1/2}.
\label{eq:dispersion}
\end{equation}
The higher dispersion of Q weights suggests that the query projection carries the most layer-specific variation, making it the natural candidate for adaptation among Q and K.

\rev{The gauge-invariance argument above concerns the product $W_Q W_K^\top$ under \emph{unconstrained} perturbations; under CRAFT's constrained parametrization the redundancy is only partial, which is consistent with the observation that adapting all four projections (CRAFT$^\star$) yields a further accuracy gain in Table~\ref{tab:commonsense_results}.}

\subsection{Design comparison of tensor-based PEFT methods}

\begin{table}[ht]
\centering
\small
\begin{tabular}{@{}lcccc@{}}
\toprule
\textbf{Property}\! & \! \textbf{LoTR} \! & \! \textbf{SuperLoRA} \! & \! \textbf{PiSSA} \! & \! \textbf{CRAFT} \\
\midrule
Decomposes & $\Delta W$ & $\Delta W$ & $W$ & $W$ \\
Cross-layer & Yes & Yes & No & Yes \\
Tucker type & Tucker-2 & Tucker-$n$ & N/A & Tucker-3 \\
What trains & Core slices & Factors & $A,B$ &    $J^{(1)}\!,J^{(2)}\!,J^{(3)}$ \\
Factors frozen & No & No & No & \textbf{Yes} \\
Residual & No & No & Yes & \textbf{Yes} \\
Indep.\ $d, N_L$ & No & No & No & \textbf{Yes} \\
\bottomrule
\end{tabular}
\caption{Design comparison of tensor-based PEFT methods. ``Indep.\ $d, N_L$'' indicates whether the Tucker adaptation parameter count is independent of model dimension $d$ and number of layers $N_L$ at fixed ranks.}
\label{tab:method_comparison}
\end{table}

\subsection{PEFT method taxonomy}

\begin{figure}[ht]
\centering
\begin{tikzpicture} 
    \draw[thick, ->] (0,0) -- (8.5,0) node[right, font=\small] {Cross-layer};
    \draw[thick, ->] (0,0) -- (0,7) node[above, font=\small, rotate=90, anchor=south] {Weight Decomp.};
    
    \draw[gray, dashed] (0,3.5) -- (8,3.5);
    \draw[gray, dashed] (4,0) -- (4,6.5);
    
    \node[font=\small, gray] at (2,0.3) {Per-layer};
    \node[font=\small, gray] at (6,0.3) {Cross-layer};
    \node[font=\small, gray, rotate=90] at (0.3,2) {$\Delta W$};
    \node[font=\small, gray, rotate=90] at (0.3,5) {$W$};
    
    \fill[craftblue!60] (1.5,1.5) circle (0.4);
    \node[font=\small\bfseries, craftblue] at (1.5,0.9) {LoRA};
    
    \fill[craftgreen!60] (2,5) circle (0.4);
    \node[font=\small\bfseries, craftgreen] at (2,4.4) {PiSSA};
    
    \fill[craftorange!80] (5.5,2) circle (0.4);
    \node[font=\small\bfseries, craftorange] at (5.5,1.4) {LoTR};
    
    \fill[craftpurple!60] (6.5,2.5) circle (0.4);
    \node[font=\small\bfseries, craftpurple] at (6.5,1.9) {SuperLoRA};
    
    \fill[gray!60] (5,1.5) circle (0.35);
    \node[font=\small\bfseries, gray] at (5,0.9) {VeRA};
    
    \fill[black] (6,5.5) circle (0.5);
    \node[font=\small\bfseries, white] at (6,5.5) {CRAFT};
    
    \draw[thick, black, ->] (3,4) -- (5.4,5.3);
    \node[font=\small\bfseries, black, align=center] at (3.5,4.8) {bridges\\both};
    
    \fill[white, rounded corners=2pt] (0.2,5.8) rectangle (4.8,7.0);
    \draw[thick] (0.2,5.8) rectangle (4.8,7.0);
    \node[font=\small, align=center] at (2.5,6.65) {\textbf{CRAFT}: Only method combining};
    \node[font=\small, align=center] at (2.5,6.25) {cross-layer + weight decomposition};
\end{tikzpicture}
\caption{PEFT method taxonomy. CRAFT uniquely combines cross-layer tensor structure with pre-trained weight decomposition\rev{, among the methods compared here}.}
\label{fig:taxonomy}
\end{figure}

\subsection{Parameter scaling analysis}
Figure~\ref{fig:param_scaling} illustrates the parameter-scaling behavior described in Proposition~\ref{prop:params_and_comparison}.

\begin{figure}[ht]
\centering
\begin{tikzpicture} 
\begin{axis}[
    xlabel={Number of Layers ($N_L$)},
    ylabel={Trainable Parameters (K)},
    xmin=10, xmax=100,
    ymin=0, ymax=1800,
    legend pos=north west,
    legend style={font=\small},
    tick label style={font=\small},
    label style={font=\small},
    grid=major,
    width=7cm,
    height=5cm,
    title style={font=\small},
    title={Parameter Count vs.\ Model Depth}
]
\addplot[thick, craftblue, mark=square] coordinates {
    (12, 196) (24, 393) (48, 786) (72, 1179) (96, 1572)
};
\addlegendentry{LoRA (r=8)}

\addplot[thick, craftgreen, mark=triangle, dashed] coordinates {
    (12, 196) (24, 393) (48, 786) (72, 1179) (96, 1572)
};
\addlegendentry{PiSSA (r=8)}

\addplot[thick, craftred, mark=*, line width=1.5pt] coordinates {
    (12, 40) (24, 41) (48, 41) (72, 41) (96, 41)
};
\addlegendentry{CRAFT (ours)}

\end{axis}
\end{tikzpicture}
\caption{Parameter count vs.\ model depth. At fixed Tucker ranks, CRAFT's Tucker adaptation parameter count remains \textbf{constant} regardless of model depth, while LoRA and PiSSA scale linearly with $N_L$. \rev{The LoRA/PiSSA curves assume $d = 1024$, rank $r = 8$ and one adapted projection per layer ($2rd = 16{,}384$ parameters per layer); the CRAFT curve corresponds to ranks} \fixw{$(\min(N_L,24),100,100)$ on two projection types, i.e.\ $40$K at $N_L = 12$ and $41$K for $N_L \geq 24$, since the mode-1 rank is capped by $r_1 \leq N_L$}\rev{. Counting LoRA on two projections would double its slope, so the comparison shown is conservative.} Whether the same ranks suffice for significantly deeper models is an open question (see Section~\ref{sec:discussion}).}
\label{fig:param_scaling}
\end{figure}

\subsection{PCA analysis of attention weights}
\begin{figure*}[t]
\centering
\includegraphics[width=\textwidth]{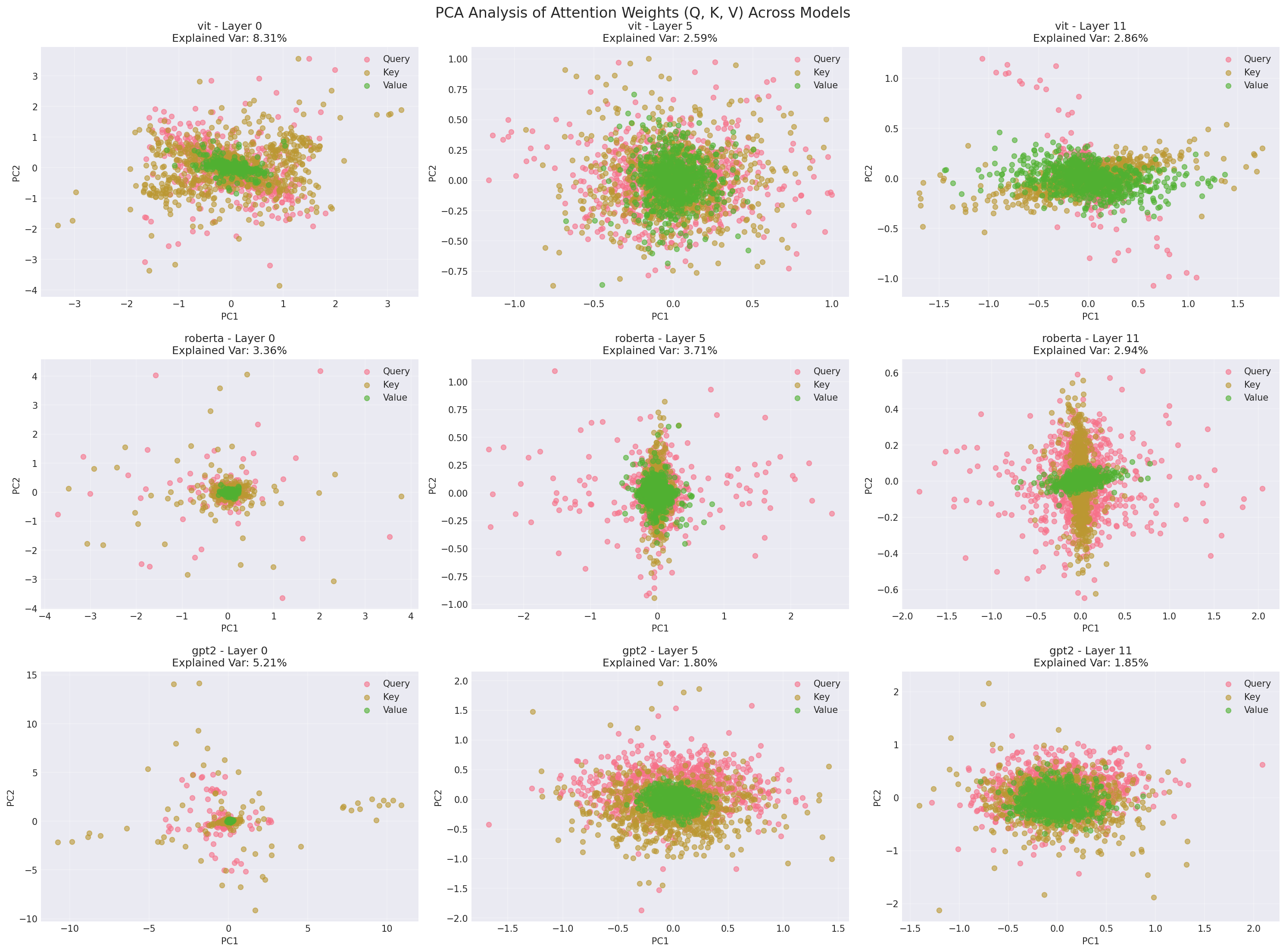}
\caption{PCA of vectorized attention weights across models and layers. Each point represents one row of a weight matrix $W_\alpha^{(\ell)}$, projected onto the first two principal components computed from the pooled set $\{w_\alpha^{(\ell)}\}$ at each layer. Q~(pink) exhibits higher dispersion (Eq.~\ref{eq:dispersion}), while K~(olive) and V~(green) concentrate near the origin. The two-component explained-variance ratio (shown in titles) decreases in deeper layers for ViT and GPT-2, indicating that the weight distribution spreads over a higher effective dimensionality.}
\label{fig:pca_analysis}
\end{figure*}

Figure~\ref{fig:pca_analysis} visualizes the attention-weight PCA analysis used to motivate the choice of adapted projections.

\section{Proofs}
\label{app:proofs}


\subsection{Proof of  Proposition~\ref{prop:params_and_comparison}}

For the sake of justification of the parameter counts above, observe that full fine-tuning trains all 
$N_L$ weight matrices of size $d \times d$, giving $O(N_L d^2)$. 
LoRA and PiSSA introduce two rank-$r$ factors per layer of sizes $d \times r$ and $r \times d$, yielding $O(N_L r d)$
parameters. LoTR shares factor matrices of size $d \times r$ across layers but maintains per-layer core slices of size $r \times r$, 
giving a parameter count of $O(N_L r^2 + rd)$. 
 
For each projection type $\alpha$, CRAFT's trainable parameters are $J^{(n)}_\alpha \in \R^{r_n \times r_n}$,
$n=1,2,3$, which results in $n_p(r_1^2 + r_2^2 + r_3^2)$ parameters independent of $d$ and $N_L$.
\rev{Note that the constraint $r_1 \leq N_L$ must hold for the mode-1 factor to be well defined; the count is independent of $N_L$ for any fixed admissible $r_1$.}
\\

\subsection{Proof of Theorem~\ref{th:init}}

Observe that 
$$
\fnorm{\mathcal{T}_\alpha^{\text{ini}} - \mathcal{R}_\alpha} = 
\fnorm{\mathcal{G}_\alpha^{\text{ini}} - \mathcal{G}_\alpha} = 
\|g_\alpha^{\text{ini}} - g_\alpha\|,
$$
\rev{where the first equality uses that the factor matrices $U^{(n)}_\alpha$ have orthonormal columns, and}
where $g_\alpha^{\text{ini}}, g_\alpha \in  \R^{r_1 r_2 r_3}$ are the vectorized versions of the tensors
$\mathcal{G}_\alpha^{\text{ini}},  \mathcal{G}_\alpha$. 
Note that, by \eqref{eq:G_alpha_ini}, one has
$$
g_\alpha^{\text{ini}} = (J^{(3)}\otimes J^{(2)} \otimes J^{(1)})\, g_\alpha.
$$
Then, since $\|g_\alpha\| = \fnorm{\mathcal{G}_\alpha} = \fnorm{\mathcal{R}_\alpha}$, we derive 
\begin{equation}
\norm{g_\alpha^{\text{ini}} - g_\alpha} \leq C(r,\epsilon, \delta) \, \fnorm{\mathcal{R}_\alpha},    
\end{equation}
where 
$$
C(r,\epsilon, \delta) = \sigma_{\max} \left( J^{(3)}\otimes J^{(2)} \otimes J^{(1)} - I_{r_1} \otimes I_{r_2} \otimes I_{r_3} \right).
$$
Here, $\sigma_{\max}(A)$ denotes the largest singular value of a matrix $A$.
\fixv{Substituting $J^{(n)} = I_{r_n} + \epsilon E^{(n)}$ and expanding the Kronecker product gives}

$$
J^{(3)}\otimes J^{(2)} \otimes J^{(1)} - I_{r_3} \otimes I_{r_2} \otimes I_{r_1}
= \epsilon \sum_{n=1}^{3} A_n + \epsilon^2 \sum_{1 \leq m < n \leq 3} A_m A_n + \epsilon^3 A_1 A_2 A_3,
$$
where $A_1 = I_{r_3} \otimes I_{r_2} \otimes E^{(1)}$, $A_2 = I_{r_3} \otimes E^{(2)} \otimes I_{r_1}$ and $A_3 = E^{(3)} \otimes I_{r_2} \otimes I_{r_1}$ all act on $\R^{r_1 r_2 r_3}$. Since $\sigma_{\max}(A \otimes B) = \sigma_{\max}(A)\,\sigma_{\max}(B)$, one has $\sigma_{\max}(A_n) = \sigma_{\max}(E^{(n)})$, and the triangle inequality together with submultiplicativity yields
\begin{align}  \label{eq:C-eps-del-bound}
  C(r,\epsilon, \delta) & \leq \epsilon \bigl[ \sigma_{\max}(E^{(1)}) + \sigma_{\max}(E^{(2)}) + \sigma_{\max}(E^{(3)}) \bigr] \\
& +  \epsilon^2 \bigl[ \sigma_{\max}(E^{(1)})\,\sigma_{\max}(E^{(2)})
+ \sigma_{\max}(E^{(1)})\,\sigma_{\max}(E^{(3)}) + \sigma_{\max}(E^{(2)})\,\sigma_{\max}(E^{(3)}) \bigr] \nonumber \\
& +  \epsilon^3 \, \sigma_{\max}(E^{(1)})\,\sigma_{\max}(E^{(2)})\,\sigma_{\max}(E^{(3)}). \nonumber
\end{align}

It is well known \citep{Tropp_2011} that, for any $\delta \in (0,1)$
and $k=1,2,3$,
\begin{equation}  \label{eq:probab_ineq}
\PP \left\{  \sigma_{\max}(E^{(k)}) \leq 2 \sqrt{r_k} +   \fixw{\sqrt{2\log(\delta^{-1})}}  \right\}  \geq 1 - \delta.
\end{equation}
Combining \eqref{eq:C-eps-del-bound} and \eqref{eq:probab_ineq} with the union bound yields
\begin{align}  \label{eq:Crepsdel_app}
& C(r,\epsilon, \delta)   \leq 3\, \left[ \epsilon \left( 2\sqrt{r} +  \fixw{\sqrt{2\log(\delta^{-1})}} \right)  \right] \\
& + 3\, \left[ \epsilon \left( 2\sqrt{r} +  \fixw{\sqrt{2\log(\delta^{-1})}} \right)  \right]^2 +
\left[ \epsilon \left( 2\sqrt{r} +  \fixw{\sqrt{2\log(\delta^{-1})}} \right)  \right]^3. \nonumber
\end{align}
Hence, if $\epsilon \left( 2\sqrt{r} +  \fixw{\sqrt{2\log(\delta^{-1})}} \right) <1$, then,
with probability at least $1 - 8 \delta$,
\begin{equation}  \label{eq:R_ini_err_more_app}
\fnorm{\mathcal{T}_\alpha^{\text{ini}} - \mathcal{R}_\alpha} \leq  7 \, 
\epsilon \left( 2\sqrt{r} +  \fixw{\sqrt{2\log(\delta^{-1})}} \right) \,\fnorm{\mathcal{R}_\alpha},
\end{equation}
so that the adapted model is initialized in close proximity to the pre-trained solution.
\rev{For instance, with $\epsilon = 10^{-3}$, $r = 128$ and $\delta = 10^{-3}$, the right-hand side of \eqref{eq:R_ini_err_more_app} equals approximately $0.18\,\fnorm{\mathcal{R}_\alpha}$, and the condition $\epsilon(2\sqrt{r} + \fixw{\sqrt{2\log \delta^{-1}}}) < 1$ is comfortably satisfied.} \fixw{At the largest rank used in this paper, $r = 360$, the same bound gives $0.29\,\fnorm{\mathcal{R}_\alpha}$.}
\\

\subsection{Proof of Theorem~\ref{thm:gradient}}

By the chain rule:
\begin{equation*}
\frac{\partial \mathcal{L}}{\partial J^{(1)}_{ij}} = \sum_{k,l} \frac{\partial \mathcal{L}}{\partial (U^{(1)}_{\text{eff}})_{kl}} \frac{\partial (U^{(1)}_{\text{eff}})_{kl}}{\partial J^{(1)}_{ij}}.
\end{equation*}
Since $(U^{(1)}_{\text{eff}})_{kl} = \sum_m U^{(1)}_{km} J^{(1)}_{ml}$:
\begin{equation*}
\frac{\partial (U^{(1)}_{\text{eff}})_{kl}}{\partial J^{(1)}_{ij}} = U^{(1)}_{ki} \delta_{jl}
\end{equation*}
Therefore:
\begin{equation*}
\frac{\partial \mathcal{L}}{\partial J^{(1)}_{ij}} = \sum_k U^{(1)}_{ki} \frac{\partial \mathcal{L}}{\partial (U^{(1)}_{\text{eff}})_{kj}} = \left( U^{(1)\top} \frac{\partial \mathcal{L}}{\partial U^{(1)}_{\text{eff}}} \right)_{ij}
\end{equation*}

\end{document}